\documentclass{article}
\usepackage{ijcai22}

\usepackage{times}
\usepackage{soul}
\usepackage{url}
\usepackage[hidelinks]{hyperref}
\usepackage[utf8]{inputenc}
\usepackage[small]{caption}
\usepackage{graphicx}
\usepackage{amsmath}
\usepackage{amsthm}
\usepackage{booktabs}
\usepackage{algorithm}
\usepackage{algorithmic}
\usepackage{subfigure}
\newtheorem{example}{Example}

\newtheorem{problem}{Problem}

\title{Random Ensemble Reinforcement Learning for Traffic Signal Control}

\author{
Ruijie Qi$^1$
\and
Jianbin Huang$^2$\footnote{Contact Author}\and
He Li$^3$\and
Qinglin Tan$^4$\and
Longji Huang$^5$\And
Jiangtao Cui$^6$
\affiliations
$^{1,2,3,4,5,6}$ School of Computer Science and Technology, Xidian University, Xi'an 710126, China 
\emails
qruijie@stu.xidian.edu.cn,
jbhuang@xidian.edu.cn,
heli@xidian.edu.cn,
qltan@stu.xidian.edu.cn,
longji.huang018@gmail.com,
cuijt@xidian.edu.cn
}
\begin{document}

\maketitle

\begin{abstract}
Traffic signal control is a significant part of the construction of intelligent transportation. An efficient traffic signal control strategy can reduce traffic congestion, improve urban road traffic efficiency and facilitate people's lives. Existing reinforcement learning approaches for traffic signal control mainly focus on learning through a separate neural network. Such an independent neural network may fall into the local optimum of the training results. Worse more, the collected data can only be sampled once, so the data utilization rate is low. Therefore, we propose the Random Ensemble Double DQN Light (RELight) model. It can dynamically learn traffic signal control strategies through reinforcement learning and combine random ensemble learning to avoid falling into the local optimum to reach the optimal strategy. Moreover, we introduce the Update-To-Data (UTD) ratio to control the number of data reuses to improve the problem of low data utilization.
In addition, we have conducted sufficient experiments on synthetic data and real-world data to prove that our proposed method can achieve better traffic signal control effects than the existing optimal methods.

\end{abstract}

\section{Introduction}

Traffic congestion has always been a problem in the construction of smart cities.
Congested vehicles not only affect the traffic efficiency of urban roads but also cause great environmental pollution and waste of resources.
There needs to be a smart traffic signal control algorithm to handle the situation, increase the throughput of intersections, reduce the travel time of vehicles, and shorten the length of the vehicles queue waiting for the green light to pass.

Traditional traffic signal control methods are usually artificially set in advance. Such as pre-defined fixed-time method\cite{Webster,fixed}, actuated control method\cite{VISSIM,actuated}, adaptive control method\cite{Haddad,SILCOCK,WONG}, optimization-based control method\cite{Varaiya2013}. 
These methods require experts to design rules based on experience. However, the transportation system is dynamically changing. Fixed rules cannot effectively control traffic signal in response to real-time changes.

Recently, researchers start to investigate reinforcement learning(RL) techniques for traffic signal control.
Reinforcement learning learns through the real-time interaction between the agent and the environment, so it has better learning ability for complex dynamic environments.
Various work based on reinforcement learning technology has performed better than traditional control methods\cite{IntelliLight,El-TantawyA12,Nishi}.

Existing reinforcement learning approaches for traffic signal control mainly focus on learning through a separate neural network.
% to increase the throughput of the intersection, reduce the travel time of vehicles, and reduce the length of wait vehicles at intersections.
Such an independent neural network may cause non-convergence, instability, or even fall into the local optimum of the training results.
%which directly affects the results of traffic signal control and causes irreparable losses.
%提出第一个问题
% Some of the latest researches in the field of reinforcement learning are about improving data utilization. The model-based method achieves higher sampling efficiency than the previous model-free method.
In the process of reinforcement learning, research on improving data utilization has also received a lot of attention in recent years. 
%The model-based method\cite{mbpo}
%achieves higher sampling efficiency than the previous model-free method. The model-based methods 
%obtain higher sampling efficiency by increasing the Update-To-Data(UTD) ratio, which is the amount of data taken by the agent to update compared to the amount of data that the agent interacts with the environment.
REDQ\cite{REDQ} introduces the Update-To-Data (UTD) ratio to improve data utilization, but it is not applicable for traffic signal control problems with discrete action spaces.
%For the model-free method, REDQ\cite{REDQ} introduce the Update-To-Data(UTD) ratio propesed by \cite{mbpo}, but this method cannot apply to the traffic signal control problem in discrete action space.
%that can improve the efficiency of data sampling
%, but this method is 
%based on the SAC algorithm and is 
%suitable for continuous motion control application scenarios. 
%提出第二个问题
%In order to solve these problems, an intuitive improvement method is to mix different single deep learning networks based on the idea of ensemble learning \cite{ensemble}to 
%learn an agent with better performance.Random ensemble reinforcement learning can better 
%stabilize the learning process and improve the learning effect.

%To solve these problems, an intuitive idea is to combine ensemble learning\cite{ensemble} and reinforcement learning by mixing multiple single deep reinforcement learning networks to stabilize the learning process, avoid falling into the local optimum, and improve the learning effect.
To solve these problems, we first randomly initialize multiple DDQN networks to conduct the learning process together according to the idea of ensemble learning\cite{ensemble}. By mixing networks, it can effectively avoid falling into the local optimum and learn the optimal traffic signal control strategy. During the interaction, all networks vote together to participate in the choice of action. During the learning process, a subset of networks is randomly selected as the target network to update all networks.
The strategies learned by multiple networks can effectively expand the strategy space, and only a subset of them used for updating can speed up the learning speed of the agent.
%To solve these problems, we first introduced ensemble learning , through which we can randomly initialize multiple deep reinforcement learning networks to learn different strategies.
%We randomly initialize multiple double DQN networks and then put the same sampling data in each network.

In this article, we propose a novel traffic signal control framework named Random Ensemble Double DQN Light(RELight), which uses a model-free algorithm for discrete action space but effectively improves data utilization.
To improve data utilization, we introduce the Update-To-Data (UTD) ratio in the traffic signal control problem, which is the amount of data taken by the agent to update compared to the amount of data that the agent interacts with the environment.
We use the Q network to score each action to select the action and solve the problem of discrete action space in traffic signal control.
We propose to use the variance of queue length as a part of the reward function.
Because the minimization of the variance of the queue length can balance the queue length in different directions at the intersection.
In this way, the learned strategy will not fall into the local optimum and avoid sacrificing waiting vehicles in one direction.
%This article extends this method to the environment of discrete action control.
%In the problem of traffic signal control, the action set is usually composed of discrete actions.
%When the agent interacts with the environment, it selects actions in the action set.
%The result shows that RELight does better than the performance of current state-of-the-art algorithms in traffic signal control.
%The result shows that RELight does better than the performance of current state-of-the-art algorithms in traffic signal control.
%Moreover, RELight can achieve this performance using fewer parameters, and with less wall-clock run time.
%There is another carefully integrated ingredient:  in-target minimization across a random subset of Q functions from the ensemble.
 
Through carefully designed experiments, we provide a specific detailed analysis on RELight.
The experimental results prove that our method can learn a more optimal strategy to control traffic signals, shorten vehicle traffic time and reduce the length of queuing vehicles. In addition, the learning process is more stable, and the data utilization rate is significantly improved.
%Our results show that using ensemble with in-target minimization, the std of the Q function can be approached to zero during the training process.
%In addition, by adjusting the number of randomly selected Q functions to minimize within the target, RELight can control the average Q function deviation. Compared with the standard integrated average and DQN with higher UTD, RELight has a lower Q function deviation standard, while maintaining a negative but close to zero average deviation in most training, thereby significantly improving learning performance.
We conducted an ablation study and the results showed that RELight is very robust in the choice of hyperparameters.

%In short, our main contributions can be summarized as follows:
%\begin{itemize}
%%	\item To improve the stability of traffic signal control, we are the first to apply random ensemble reinforcement learning for traffic signal control.
%	\item We propose a novel model named RELight to apply random ensemble reinforcement learning for traffic signal control .
%	\item We introduced UTD ratio to effectively enhance data utilization efficiency for traffic signal control.
%	\item %Through comprehensive experiments on synthetic traffic data and real-world traffic data, we prove that the control strategy learned by RELight is better than the existing RL methods.
%	We propose to use the variance of queue length as a part of the reward function.
%	% It is because the variance of the queue length can balance the queue lengths in different directions at the intersection well so that the learned strategy will not fall into the local optimum and avoid the sacrifice of waiting vehicles in one direction.
%	
%\end{itemize}

The rest of this paper is organized as follows. Section 2 discusses the literature.
Section 3 formally defines the problem. The method is shown in Section 4 and the experiment results are shown in Section 5. Finally, we conclude the paper in Section 6.

\section{RELATED WORK}
In this section, we firstly introduce traditional traffic signal control methods, then introduce reinforcement learning methods\cite{SuttonB98}.

\subsection {Conventional Traffic Signal Control}
% {\bfseries Traditional traffic signal control.} 
Many traffic signal control methods were first proposed in the field of transportation research.
%Existing methods can generally be classified into the following four categories:

Pre-defined fixed-time control method\cite{offset,Traffic} is a control method that is completely designed in advance by human experts based on experience.
%This method is the simplest. 
%It is easy to implement but lacks real-time control of dynamic traffic conditions. 
%Not only is it possible to cause traffic congestion, but it is also not suitable for the development of today's smart cities.
Actuated control is method also composed of pre-defined rules, but traffic signals can be adapted to real-time traffic data. 
%It can adjust traffic signals according to the rules defined for traffic information at intersections.
%For example, when the queue length in one direction of an intersection reaches the threshold in the rule, the traffic signal can change for the vehicles in this direction, and the waiting vehicles in this direction will be released.
Adaptive control method is not a set of rules but a set of traffic signal plans. 
%The most suitable traffic signal will be selected according to the current traffic conditions. 
%This method is the most widely used in traffic transportation today. 
%For example, it has been adopted in traffic signal control systems, such as SCATS\cite{SCATS,SCATS2} and SCOOT\cite{SCATS,SCOOT}.
Optimization-based control method\cite{TRANSYT} can be adjusted according to real-time traffic conditions, unlike the previous method that relies on some hand-designed external rules.
%This method formalizes the traffic signal control problem as an optimization problem.
%They set a traffic flow mode based on the actual traffic situation and a large number of assumptions, and then optimize the traffic signal control in this mode.
%The phase ratio and cycle period are assumed based on the traffic flow patterns observed from the traffic data.
Although this method is based on traffic data for traffic signal control, it uses too many assumptions and is difficult to apply in practice.

\subsection {Reinforcement Learning for Traffic Signal Control}

Recently, many works have applied reinforcement learning technology to the research of traffic signal control.
These results show that reinforcement learning has a better effect on the control of dynamic systems. 
%Moreover, since it learns through the temporary feedback from the environment, it does not require many complex pre-definitions.
Q-learning is the most classic algorithm in reinforcement learning algorithms. 
It is suitable for scenarios in discrete action space, and the action space in traffic signal control problems is conventionally defined as a set of discrete actions\cite{IntelliLight}.
Some work has also achieved good results by redefining the traffic state and also using the DQN network model\cite{PressLight}.
There is also work by redefining the phase of the traffic signal and constructing the corresponding network structure for effective learning and control\cite{FRAP}.

However, Q-learning may usually cause over-estimation of the Q value.
Due to the instability of the dynamic environment and the random initialization of the deep reinforcement learning network, the final convergence process of the reinforcement learning algorithm is prone to be unstable, and large deviations occur during different training processes. 
In order to improve this situation, we naturally thought of the ensemble idea of machine learning.

Research on the characteristics of ensemble learning\cite{ensemble}, uses some simple model designs to analyze different levels of the ensemble.
Combining ensemble learning with reinforcement learning,\cite{REDQ}has given a good demonstration.
%Randomly ensemble N seperate DQN models, and by separately establishing the target network and M of them are randomly selected for the entire network update. 
%There are still not many ways to use ensemble learning for reinforcement learning.
%In this paper, for the first time, random ensemble learning and deep reinforcement learning are applied to traffic light control problems, and have achieved results that exceed the existing optimal algorithms.
In this paper, for the first time, random ensemble learning and deep reinforcement learning are applied to traffic light control problems and have achieved good results.

\section{Preliminary}
\subsection{Environment}
In this paper, we study traffic signal control in a single intersection scenario.
We use a standard four-way intersection as an example for the following explanation. 
But these concepts can be easily transplanted to a three-way or a five-way intersection.

\begin{itemize}
	\item \textbf{Entering approach}: The four directions of each intersection are named: North, South, East and West('N','S','E','W' for short).
	\item \textbf{Signal phase}:
	% A signal light can generally control the flow of traffic in three directions in the current lane: Left, Through and Right('L','T','R' for short).  
	%A signal phase that can effectively control traffic flow should have a couple of corresponding directions that are green lights, and other directions are red lights. 
	In order to define the problem in a simple and formal way, we assume that the traffic signal has only two phases: a green light in the east-west direction and a red light in the other direction(WEG); a green light in the north-south direction and a red light in the other direction(NSG). 
	%The green light includes T and turning L, and R can pass through by default according to actual conditions.
	%The yellow light is used to clear the vehicles that have entered the intersection when the green light is switched to the red light.
\end{itemize}

\begin{table} 
	\centering 
	\begin{tabular}{cl} 
		\toprule 
		Notation  & Meaning \\ 
		\midrule 
		$S$        & state space      \\
		$A$        & action space       \\ 
		$R$        & reward function      \\ 
		$\gamma$        & discount factor      \\
		$L$        & queue length      \\ 
		$\bar L$ & average value of queue length on all lanes\\
		$l_i$    & queue length on each lane i   \\
		$w_i$    & vehicle waitong time on each lane i   \\
		$n_i$    & number of vehicles in the queue on each lane i   \\
		$p_c$    & current phase   \\
		$p_n$    & next phase   \\
		$V$      & variance of queue length   \\
		$d_i$    & Delay of vehicles on each lane i   \\
		$w_i$    &  waiting time of vehicles on each lane i   \\
		$C$    & whether to switch phase signal   \\
		$N$    & number of vehicles passing through the interse   \\
		       & -ction in a time step \\
		$T$    & travel time of vehicles N passing in a time step  \\
		\bottomrule 
	\end{tabular} 
\caption{Notation} 
\label{tab:booktabs} 
\end{table}

\subsection{Problem Defination}
%Based on the real-time feedback of the dynamic traffic system for learning, we propose a method for traffic signal control through reinforcement learning.
In our problem, an intersection is controlled by an independent agent.
The agent observes the traffic environment, makes actions, and then learns according to the rewards of the environment's feedback.
The goal of the agent is to learn an optimal strategy for manipulating traffic phases in order to optimize the travel time of the vehicle through the intersection.
This traffic signal control problem can be formalized as a Markov decision process $<S,A,P,R,\gamma>$\cite{SuttonB98}.

\begin{problem}
	Given the state space $S$, the action space $A$, and a reward function $R(s,a)$. 
	Our work is based on the model-free method, so the state transition probability function $P$ is unknown.
	The goal of the agent is to learn an optimal strategy $\pi(a|s)$ to maximize the expectation of discounted rewards.
	The agent selects best actions for different states according to this strategy. 
\begin{align} 		 		 	
	G_t =& R_{t+1}  + \gamma R_{t+2} + \gamma^2 R_{t+3} + ... \nonumber\\
	=& \sum_{k=0}^\infty \gamma^k R_{t+k+1}.   	 
\end{align}
	
\end{problem}

%The detailed definitions of state space, action space, and reward function are as follows:
\section{METHOD}

%Solving the problem of traffic light control through reinforcement learning has gradually become a hot issue in recent years.
%With the widespread application of deep learning, most of the existing problem-solving ideas are to adopt deep reinforcement learning.
%But most of the solutions still remain in the use of simple and separate deep reinforcement learning models for learning.
%There are many parameters in the deep learning network, and there is a strong randomness, which may produce very different results in different situations.
%A single DQN is prone to overestimation problems in the learning process.
%Therefore, it is intuitive to think of using ensemble learning to solve this problem.
%In addition, in order to improve data utilization, we are inspired by model-based methods.
%There have been works using UTD parameters in continuous action space tasks. In this paper, we extend it to discrete action space tasks.

%In this section, we will mainly introduce the design of the entire RElight framework.
%First, we introduce the basic algorithm DDQN used by the RELight framework, and then introduce ensemble learning, and combine it with DDQN to form the random ensemble Light.
%Finally, the UTD ratio is introduced and its application in discrete action space is studied.

\subsection{Framework}

In order to reduce the variance in the Q function estimation, we adopt the ensemble learning framework REDQ as our algorithmic framework\cite{REDQ}.
%Our method can directly interact with the traffic environment for signal light control through online learning, and achieve good results without complicated assumptions and pre-training.
Similar with REDQ, standard reinforcement learning is divided into two parts: acting and learning.
The RELight framework we proposed is shown in Figure 1. The dotted line in Figure 1 is acting, and the solid line is learning.
\begin{figure}[ht]
	\centering
	\includegraphics[width=1.0\linewidth]{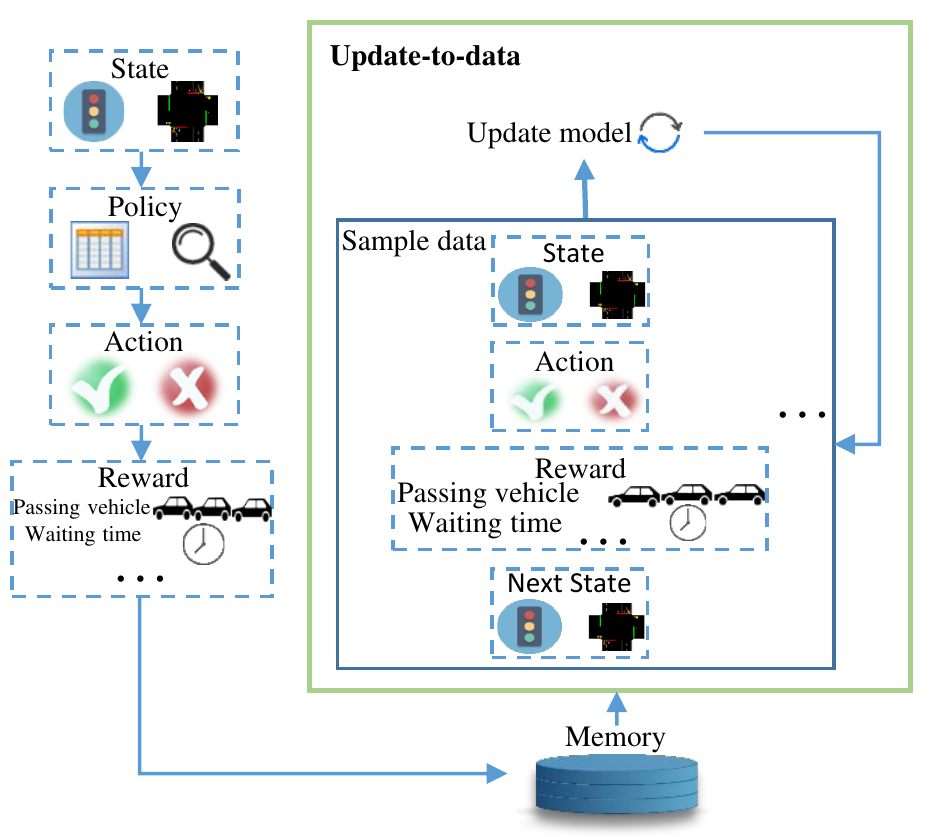}
	\caption{Model framework.}
\end{figure} 
%The blue part in the figure 1 is acting, and the green part is learning.

First, the acting part selects the actions based on the maximum Q value of each Q network or exploration and then votes on the actions group to determine the action to perform.
Then the state of the environment, the action performed, and the rewards for environmental feedback will be deposited in the memory.
The learning part samples training data from memory to update the model.
In the process of model update, due to ensemble learning, multiple Q functions need to be minimized.
However, instead of minimizing all Q functions in the target, we randomly select some of them to minimize.
More importantly, the two parts can be performed at the same time, and the data sampling is used to update the model while generating the training data.
The framework of this algorithm can effectively control the deviation of overestimation and the variance of Q value estimation.

\subsection{Agent Design}
In this part, we introduce the specific design of state space, action space, and reward function.
\begin{itemize} 
	\item State space $S$.
	In this paper, we define the state of one intersection.   
	For each lane $i$ in the intersection, the state consists of the following parts: queue length $l_i$, vehicle waiting time $w_i$, and the number of vehicles in the queue $n_i$, and traffic signal state information: current phase $p_c$, next phase $p_n$.
	
	\item Set of actions $A$.
	The action space is defined as whether to change the phase of the traffic signal $a$.
	If the agent decides to switch traffic signal phase, set $a=1$; otherwise, $a=0$.
	
	%	\item Transition probability $P$.
	%	In this paper, we focus on model-free methods, and the state transition probability matrix is unknown, which belongs to the definition in model-based methods.
	
	\item Reward $R$.
	In this paper, reward is defined as the weighted sum of the following parts:
	
	\begin{align} 		 			
		Reward =& w_1*V + w_2*\sum_{i \in l}d_i + w_3*\sum_{i \in l}w_i \nonumber\\
		+& w_4* C + w_5* N + w_6* T. 	
	\end{align}
	
	The detailed definition of each item in the reward function is as follows:
	\begin{enumerate}            
		\item The variance of queue length on all approching lanes.
		The queue length is defined as the number of waiting vehicles on each road, 
		%and vehicles with a speed lower than 0.1m/s are classified as waiting vehicles.
		we count the number of waiting vehicles on each road, and then calculate their variance on all lanes.
		\begin{equation} 		 
			\begin{split} 		 	
				V = \overline{\sum_{i \in l}(l_i - \bar L)^{2}}.
			\end{split}   	 
		\end{equation}
		
		\item Delay of each lane i.            
		\item Waiting time of vehicles on each lane i.    
		\item Whether to switch the sign of the traffic light, it only contains 0 or 1. 
		\item The number of vehicles N passing through the intersection in a time step.
		\item The travel time of vehicles N passing in a time step.
	\end{enumerate} 
	
\end{itemize}
\subsection{Network Structure}
%In this paper we use Q-network to estimate the action value for action selection.
%Since ensemble learning requires multiple Q networks, we vote from different action choices generated by multiple Q networks to finally determine the action to be taken at the next moment.
The specific network structure is shown in Figure 2.
In the learning process, we adopted Double DQN as the core Q network.

%Since the action space is discrete and multiple networks need to be integrated, we choose the most classic deep Q-learning network $Q(s,\cdot;\theta)$ as the basic network structrue, where $\theta$ is parameters in the network. 
Basically, our network takes the vehicle state and the traffic signal phase state at the intersection as input. 
The output is the predicted score for each action according to the Bellman equation: 
\begin{equation} 		 		 	 		
	%Q(s_t,a_t) = R(s_t,a_t)  + \gamma maxQ(s_t+1,a_t+1)	  
	y_t = r_{t+1} + \gamma \mathop{\max}_a Q(s_{t+1}, a; \theta_t).
\end{equation}

%For each separate model, we adopted the double DQN design.
We choose N Double DQN networks for ensemble to learn the optimal strategy.
%Then we randomly select M of these N networks to calculate the loss for updating the network. 
%The method of separating the prediction network  and the target network $Q(s,\cdot;\theta')$ also use to eliminates the over-estimation.
In the double DQN algorithm\cite{doubleQ}, the calculation of the Q value is no longer the calculation of the maximum Q value of each action through a separate network as, but choose the action corresponding to the maximum Q value is first found in the prediction network $Q(s,\cdot;\theta)$:

\begin{equation} 		 		 	 		
	a^{max}_{m} = \mathop{\arg\max}_{a} Q_{m}(s_{t+1},a;\theta_{t}).
\end{equation}
%We vote to determine the final action based on the actions selected in the M Q networks.
Then use the selected action $a^{max}_{m}$ to calculate the target Q value in the target network $Q^{\prime}$, and choose the min Q value among the M sub-Q networks:

\begin{equation} 		 		 	 		
	y_t = r_{t} + \gamma \min_{m\in{M}}Q^{\prime}_{m}(s_{t+1}, a^{max}_{m} ; \theta_t^\prime).
\end{equation}
Then we use the expand $Y_t$ to update all N networks.
The target network revalues the parameter from the prediction network after each update through soft update\cite{softupdate}.
%, and uses it to update its own parameters:

%\begin{equation} 		 		 	 		 	
%	\theta_{t+1}^\prime = \rho\theta_t^\prime + (1-\rho)\theta_{t}.
%\end{equation}
%Where 0 \textless\textless $\rho$ \textless 1.
%In this way, the change of the target network parameters is small, and the calculated target value changes relatively smoothly.

\begin{figure}[htb]
	\centering
	\includegraphics[width=0.85\linewidth]{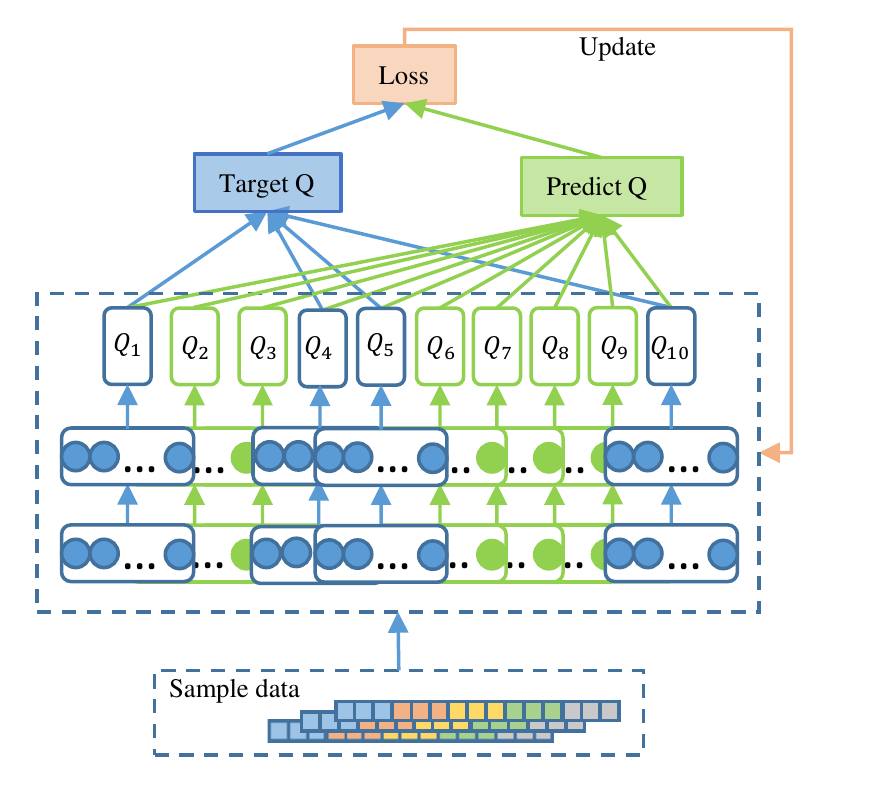}
	\caption{RElight framework.}
\end{figure} 

The traffic signal control process in the real world is very complicated. We will use Example 1 to explain it.
\begin{example}
	We assume there is an intersection with two phases, West-East direction is green (WEG) and North-South direction is green (NSG).
	We take the current intersection state and traffic signal phase as input. Each DDQN learns different strategies, and even if the observed state is the same, it is possible to score different results for each action. Therefore, each DDQN chooses to switch the traffic signal state or maintain the current signal phase according to its strategy. Then we vote in these actions to decide whether to change the signal phase.
	 
\end{example}
\subsection{update-to-data ratio}
%\paragraph{UTD ratio}
The model-based method uses the UTD ratio, which is the ratio of the amount of data the agent uses for learning to the amount of data that the agent actually interacts with the environment, which greatly improves the sampling efficiency.
This is because the model-based method can perform data simulation through the model, and mix the simulated data with the real data to improve the sampling efficiency, and the UTD ratio can reach up to 40.
%In addition, the UTD ratio was previously applied to tasks in continuous action space, but this article applies it to discrete action space scenarios for the first time.

However, the previous research on UTD ratio mainly focused on the model-based methods, and the UTD ratio based on the model-free methods is less researched. In our model learning process, we take multiple samples cyclically from memory to enhance data utilization. Moreover, we have proved that the UTD ratio can effectively improve data utilization and enhance the learning effect through synthetic data and real-world data in the experiments.	
\subsection{Algorithm}
%The entire algorithm of RELight is described in detail in Algorithm 1. G is the Update-To-Data ratio. We use the soft update method when updating the network.
%The process of traffic signal control intersection and the update process of traffic signal control strategy are shown in Algorithm 1. G is the Update-To-Data ratio. We use the soft update method when updating the network.
The process of online interaction between the agent and the intersection, the update process of the traffic signal control strategy is shown in Algorithm 1. G is the Update-To-Data ratio. We use the soft update method to update the networks.
%The process of traffic signal control and strategy update process are shown in Algorithm 1. G is the Update-To-Data ratio. We use the soft update method when updating the network.
\begin{algorithm}[h]
	\caption{Random Ensemble Double DQN Light} 
	\label{alg:algorithm} 
	\begin{algorithmic}[1] %[1] enables line numbers 
		\STATE  Initializes N Q-function parameters $\theta_i$, i=1,2,...,N, empty replay buffer $D$. Set target parameters  $\theta_i^\prime$, i=1,2,...,N	 				 			
		\WHILE {Agent is interacting with the Environment}  			
		\STATE choose $a_i$ $\in$	$A$	based on $Q_i(s,a;\theta_{t})$	 			
		\STATE Vote for the performed action $a_t$ in set $\{a_i\}_i^N$  			
		\STATE Collect state $s_{t+1}$ and reward $r_t$ from the envionment by taking action $a_t$ 			
		\STATE Store transitions $(s_t, a_t, r_t, s_{t+1})$ in replay buffer $D$ 			
		\FOR {$G$ updates} 						 			
		\STATE Sample a mini-batch $B = \{(s_t, a_t, r_t, s_{t+1})\}$ from replay buffer $D$						 			
		\STATE Randomly sample $m$ different numbers from $\{1,2,...,N\}$ as the index of choosed Q-function.						 			
		\STATE Calculate target $Y$ in the selected $Q_m, m\in M$:
		%$$Y_t = r_{t} + \gamma Q(s_{t+1}, a_t ; \theta_t^\prime).$$ 
		$$y_t = r_{t} + \gamma \min_{m\in{M}}Q^{\prime}_{m}(s_{t+1}, a^{max}_{m} ; \theta_t^\prime).$$			
		\FOR {$i=1,2,...,N$}	 			
		\STATE Update $\theta_i$ with gradient descent using 			$$\bigtriangledown_\theta\frac{1}{|B|} \sum_{(s_t, a_t, r_t, s_{t+1}) \in B} (Q_i(s,a;\theta_{t}) - y_t)^2 $$ 			
		\STATE Update target networks with $\theta_i^\prime\gets\rho\theta_i^\prime + (1-\rho)\theta_i$ 			\ENDFOR				 			
		\ENDFOR 			
		\STATE $s \gets s_{t+1}$								 			
		\ENDWHILE 
	\end{algorithmic} 
\end{algorithm}
\section{EXPERIMENT}
In this part, we conducted experiments using synthetic data and real-world data.
After that, we showed the results of comparing our method with other methods and some case studies.
\subsection{Settings}
Following the tradition of traffic signal control research, the simulation platform we used in the experiment is SUMO.
%(Simulation of Urban MObility).
%SUMO provides convenient APIs for road network design, traffic environment simulation , and traffic signal control.
%As long as the traffic flow pattern is given, the simulator will make the vehicle drive normally according to the traffic environment.
We obtain the state of the traffic environment according to the simulator as the input of the agent, and then pass the action of the agent to the simulator through the traffic light API to control the phase of the traffic light, and finally get feedback rewards from the simulator.
According to general rules, there should be a three-second yellow light between the green light and the red light, so that vehicles that have entered the intersection can safely leave the intersection.

The parameter setting in the experiment and the weight coefficient of the reward function are listed in Table 2 and Table 3 respectively.

\begin{table}[htbp]
	\centering
	\begin{tabular}{cc}
		\toprule
		Model Parameter & Value  \\
		\midrule
		 Action time interval         & 5 seconds                 \\ 
		$\gamma$ for future rewarad       &0.8            \\ 
		$\epsilon$ for exploration       & 0.05                  \\
		batch size        & 20                 \\
		memory length     & 1000                 \\
		learning rate     & 0.01             \\
		number of Q-function N     & 10                \\
		number of Q-function in-target M    & 4                 \\
		UTD ratio    & 40                \\
		model soft upadate polyak    & 0.995               \\
		\bottomrule
	\end{tabular}
	\caption{Parameter Settings}
	\label{tab:booktabs}
\end{table}

\begin{table}[htbp] 
	\centering 
	\begin{tabular}{cccccc} 
		\toprule 
		$w_1$ & $w_2$   & $w_3$  & $w_4$  & $w_5$  & $w_6$ \\
		\midrule 
		-0.25      & -0.25  & -0.25  & -5  &  1 &  1\\
		\bottomrule 
	\end{tabular} 
\caption{Weight Coefficient of the Reward Function} 
\label{tab:booktabs} 
\end{table}

\subsection{Datasets}
\subsubsection{Synthetic data}
The experimental environment for synthetic data is a four-way intersection. The road length in the four directions of this intersection is 150 meters, and there are six lanes in each direction, three entry lanes and three exit lanes.There are two phases of traffic lights: WE-Green for Through and Left(WEG), NS-Green for Through and Left(NSG). The specific data characteristics are shown in Table 4.
%It mainly includes four traffic flow conditions: 

\begin{table}[htbp] 
	\centering 
	\begin{tabular}{ccrrr} 
		\toprule 
		Config & Begin & End  & Number  & Condition \\
		 & (s) & (s)  & (cars)  &  \\
		\midrule 
		switch & 0 & 36000 & 14400 & WEG \\ & 36001 & 72000 & 14400 & NSG\\
		%switch & WE & 0.400 & 0 & 36000 \\ & SN & 0.400 & 36001 & 72000 \\
		equal & 0 & 72000 & 2400 & WEG \\ & 0 & 72000 & 2400 & NSG\\
		%equal & WE & 0.033 & 0 & 72000 \\ 		& SN & 0.033 & 0 & 72000 \\
		unequal & 0 & 72000 & 14400 & WEG \\ 		& 0 & 72000 & 2400 & NSG \\
		synthetic & 0 & 72000&  & switch  \\
		& 72001 & 144000&  & equal  \\ 
		& 144001 & 216000&  &  unequal \\		
		\bottomrule 
	\end{tabular} 
\caption{Configurations for synthetic traffic data. It mainly includes four traffic flow conditions: (a) switch direction traffic flow, (b) equal traffic flow, (c) unequal traffic flow, and (d) complex traffic flow, which is a collection of the above situations.} 
\label{tab:booktabs} 
\end{table}

%\begin{table}[htbp] 
%	\centering 
%	\begin{tabular}{ccrrr} 
%		\toprule 
%		Config & Directions & Arrival rate  & Start time  & End time \\
%		&  & (cars/s)  & (s)  & (s) \\
%		\midrule 
%		switch & WE & 0.400 & 0 & 36000 \\
%		& SN & 0.400 & 36001 & 72000 \\
%		equal & WE & 0.033 & 0 & 72000 \\ 		& SN & 0.033 & 0 & 72000 \\
%		unequal & WE & 0.200 & 0 & 72000 \\ 		& SN & 0.033 & 0 & 72000 \\
%		synthetic & switch &  & 0 & 72000 \\
%		& equal &  & 72001 & 144000 \\ 
%		& unequal &  & 144001 & 216000 \\		
%		\bottomrule 
%	\end{tabular} 
%	\caption{Configurations for synthetic traffic data} 
%	\label{tab:booktabs} 
%\end{table}

\subsubsection{Real-world data}
The real-world dataset is collected by surveillance cameras in Hangzhou.
The data collect the traffic situation of each direction in every second of a four-way intersection within an hour.
By analyzing the data recorded by the camera, we obtained the trajectory data when the vehicle passed the intersection, and put it as traffic flow data into SUMO for experiment.
The details of the data set are listed in the table 5.

\begin{table}[htb] 
	\centering 
	\begin{tabular}{lrr} 
		\toprule 
		Dataset  & time(s) & Arrival rate(cars/s) \\ 
		 \midrule 
		 Hangzhou & 3600    & 0.514                \\ 
		 \bottomrule 
	 \end{tabular} 
 \caption{Details for real-world traffic data} 
 \label{tab:booktabs} 
\end{table} 
\begin{figure*}[htbp]
	\centering
	\subfigure[RELight with different UTD]{
		\includegraphics[width=5cm]{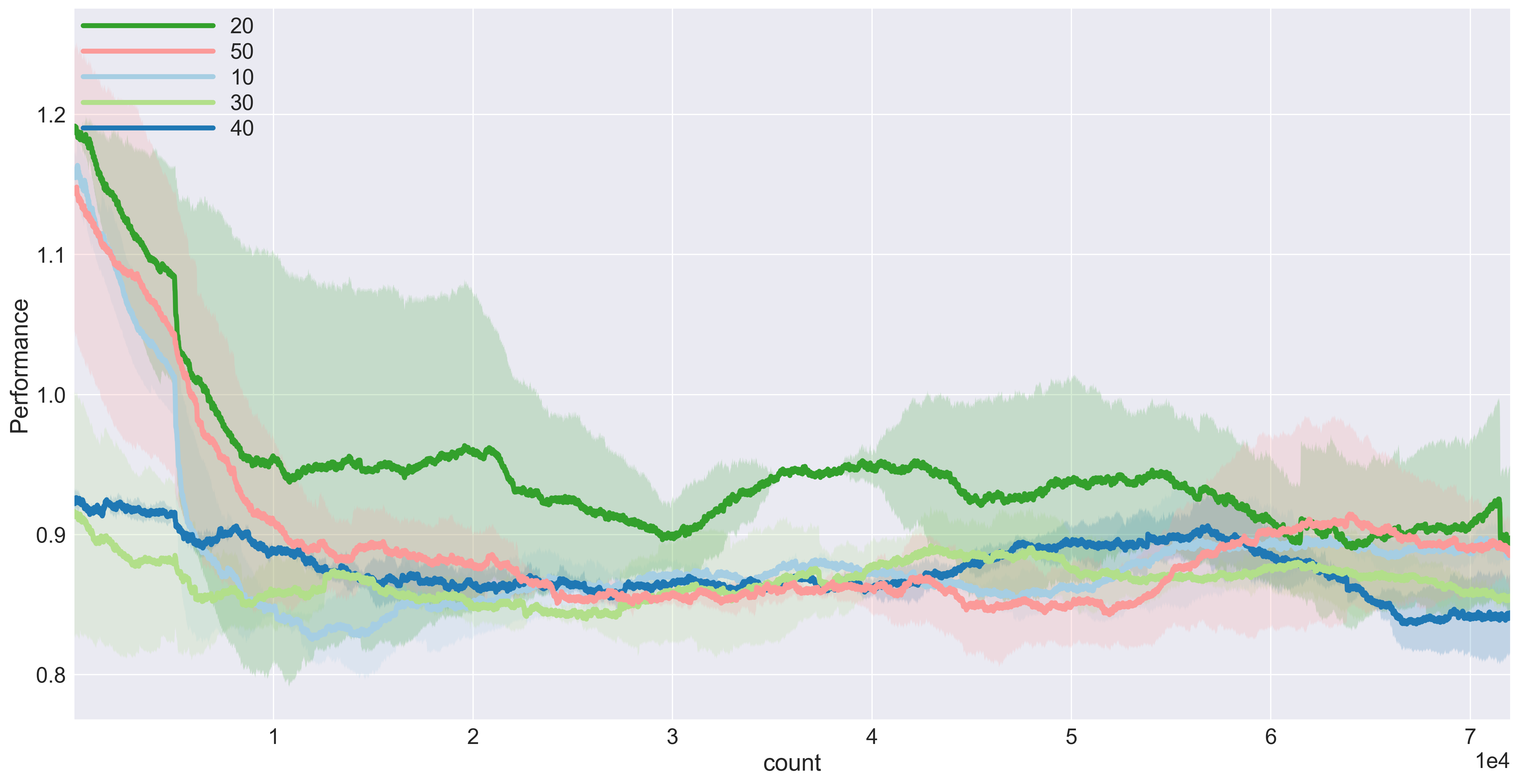}
		%\caption{fig1}
	}
	\quad
	\subfigure[RELight with different N]{
		\includegraphics[width=5cm]{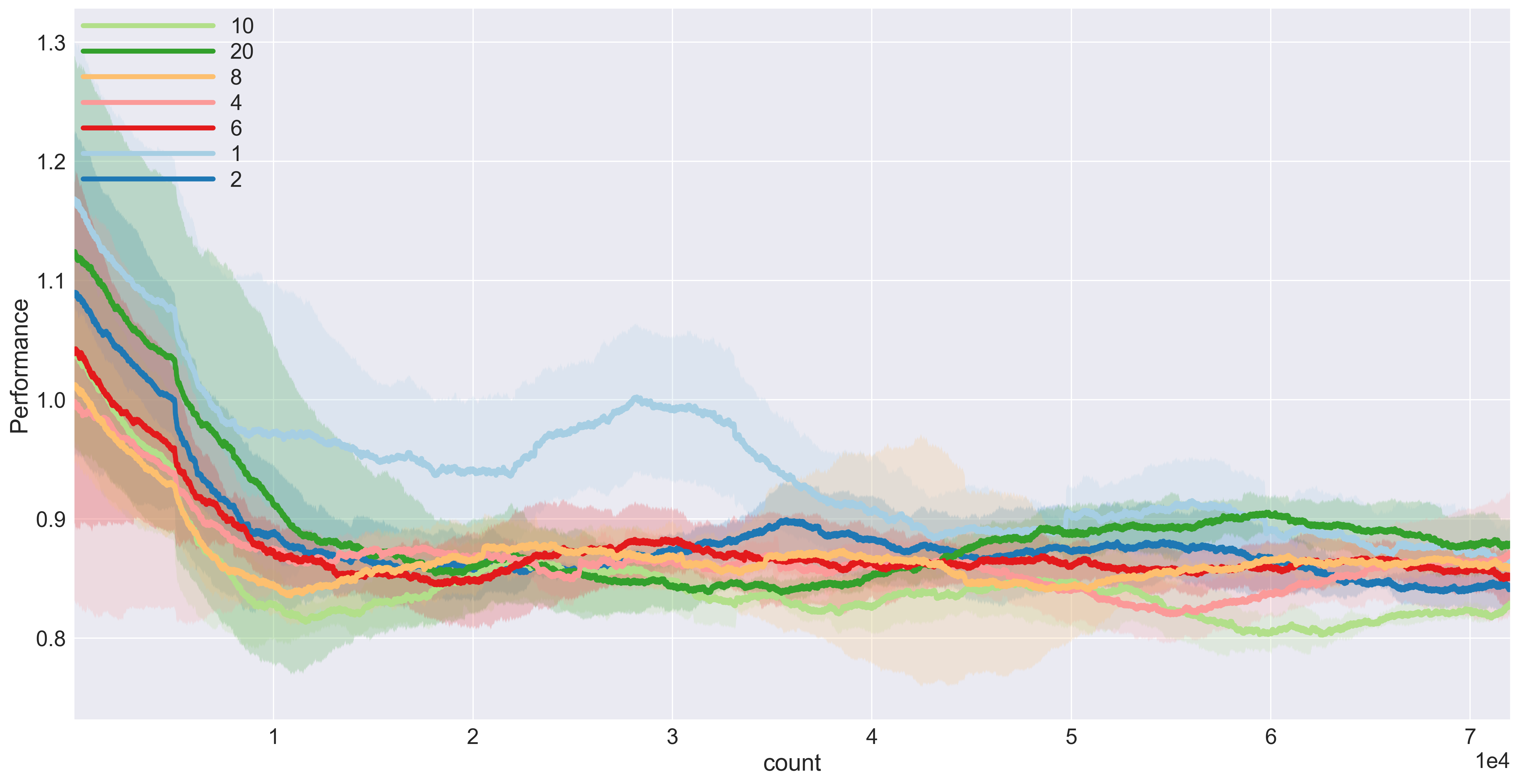}
	}
	\quad
	\subfigure[RELight with different M]{
		\includegraphics[width=5cm]{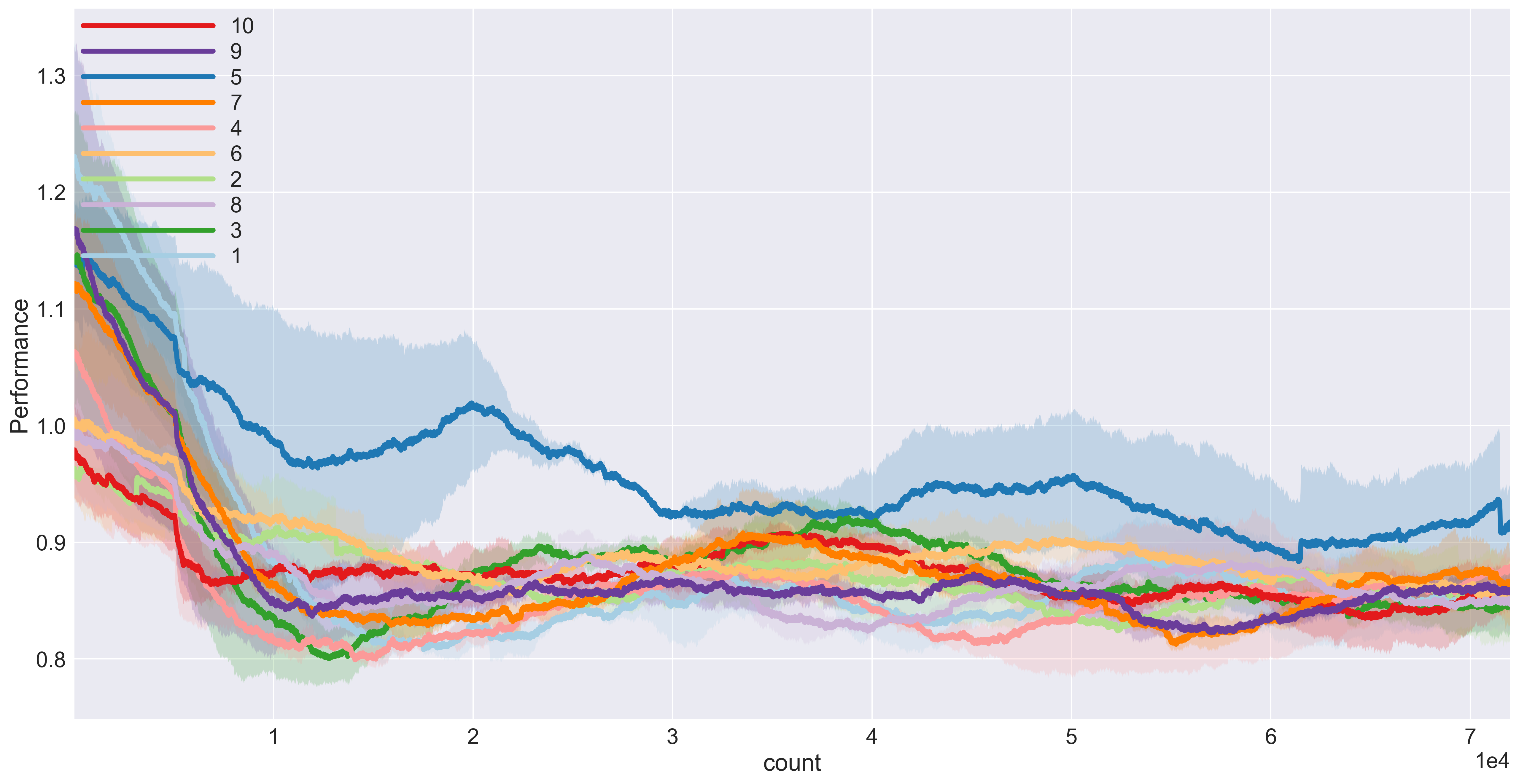}
	}
	\caption{Study of RELight}
\end{figure*}
\subsection{Preformance Comparison}
\paragraph{Compared Methods}
In order to measure the effect of our model, we used two types of methods to compare the experimental results: traditional transportation methods and RL methods.
%In order to make a fair comparison, the pre-training part is eliminated in the RL method. 
%In addition, we have carefully debugged all methods to ensure the best results.
Traditional methods include Fixed-cycle control method\cite{fixed} and Self-Organizing Traffic Light  (SOTL) control method\cite{SOTL}. RL methods include DRL\cite{DRL} method and IntelliLight\cite{IntelliLight} method.
\paragraph{Evaluation Metric}
We follow the common metrics of existing research to compare the results of different methods: The average queue length at the intersection (Queue length). The average delay at the intersection(Delay). The average travel time of a vehicle through an intersection(Travel time).
%\subsubsection{Comparison with the performance of mentioned earlier methods on synthetic data}
\paragraph{Comparison with the performance of mentioned earlier methods on synthetic data.}
We first use four baselines and our proposed method to conduct comparative experiments on four synthetic datasets.
It can be seen from the following four tables that our proposed method performs better than the baseline in the four synthetic situations, especially on the metric of travel time.
In configuration 1, the SOTL method is similar to our results in the first two metrics, and in config unequal, the IntelliLight method has a similar situation.
But our method shows more obvious advantages in more situations, which shows that our method has the ability to deal with more complex and changeable situations and the generalization ability to different situations.
RELight learned a better traffic signal control strategy through random ensemble and improvement of data utilization.

\begin{table}[ht]  	
	\centering  	
	\begin{tabular}{crrrr}  		
		\toprule  		
		Data & Model name   & Queue& Delay       & Travel time      \\
		&&length&&\\  		 
		\midrule  		 
		switch& Fixed-cycle  & 8.532        & 2.479       & 42.230        \\
		&SOTL         & 0.006        & 1.598       & 24.129        \\
		&DRL          & 91.412       & 4.483       & 277.430       \\
		&IntelliLight & 8.125  & 1.883 & 56.109 \\
		&RELight      & 0.076        & 1.619       & 3.841         \\		
		\midrule  		  		
		equal&Fixed-cycle  & 1.105        & 2.614       & 34.278        \\ 		
		&SOTL         & 19.874        & 4.384       & 177.747        \\ 		
		&DRL          & 3.405       & 3.431       & 52.075       \\ 		
		&IntelliLight & 0.261  & 1.760 & 0.868 \\ 		
		&RELight      & 0.223        & 1.733       & 0.835         \\		  		
 \midrule  		  		 		
 unequal&Fixed-cycle  & 4.159        & 3.551       & 36.893        \\ 		 		
 &SOTL         & 20.227        & 5.277       & 69.838        \\ 		 		
 &DRL          & 16.968       & 4.704       & 66.485       \\ 		 		
 &IntelliLight & 0.865  & 2.676 & 3.133 \\ 		 		
 &RELight      & 0.896        & 3.111       & 3.016         \\		
\midrule  		  		 		 		
synthetic&Fixed-cycle  & 4.601        & 2.883       & 39.707        \\ 		 		 		
&SOTL         & 13.372        & 3.753       & 54.014        \\ 		 		 		
&DRL          & 91.887       & 4.917       & 469.417       \\ 		 		 		
&IntelliLight & 3.062  & 2.177 & 18.782 \\ 		 		 		
&RELight      & 0.485        & 2.161       & 2.728         \\	 			 
		\bottomrule  	 
	\end{tabular}   
\caption{Performance on synthetic data}   
\label{tab:booktabs}  
\end{table} 

\paragraph{Comparison with the performance of mentioned earlier methods on real-world data.}
On the real-world dataset, we also compared our proposed method with the baseline method through experiments.
According to table 7, it can be clearly seen that our proposed method has obvious improvement in queue length and travel time.
The ensemble learning and multiple updating of RELight effectively avoid overestimation of the Q value and effectively reduce excessive actions.
It prevents the traffic flow in one direction at the intersection to make it wait for a long time.
Therefore, it can avoid falling into the local optimum and learn a better global optimization strategy.

\begin{table}[htb]  	 	 	 	 	
	\centering  	 	 	 	 	
	\begin{tabular}{crrr}  		 		 		 		 		
		\toprule  		 		 		 		 		
		Model name   & Queue length & Delay       & Travel time      \\  		  		 		\midrule  		  		 		 		 		
		Fixed-cycle  & 19.542        & 3.377       & 84.513        \\ 		 		 		 		
		SOTL         & 16.603        & 4.070       & 64.833        \\ 		 		 		 		
		DRL          & 54.148       & 4.209       & 166.861       \\ 		 		 		 		
		IntelliLight & 8.425  & 4.44 & 20.920 \\ 		 		 		 		
		RELight      & 2.733        & 3.610       & 5.837         \\		  		 		 		\bottomrule  	  	 	 	 	
	\end{tabular}       
\caption{Performance on real-world data}       
\label{tab:booktabs}      
\end{table}

\subsection{Study of RELight}
\paragraph {UTD}
When the value of the UTD ratio is different, the learning effect is also different. We have carried out comparative experiments on different values of UTD parameters when the other parameters are optimal(N=M=10). 
%Among them, the UTD value set in this article of REDQ\cite{REDQ} is 20.
Taking the synthetic data config unequal as the traffic condition, and comparing the UTD values with different values, the following figure is obtained.
%Based on the various indicators, it can be seen that the optimal value of UTD should be 40.
The introduction of UTD effectively improves the stability of the learning process. When UTD=40, the agent can learn the optimal strategy.

%\begin{table}[htb] 	 	 	 	 	 	
%	\centering  	 	 	 	 	 	
%	\begin{tabular}{crrr}  		 		 		 		 		 		\toprule  		 		 		 		 		 		
%		UTD   & Queue length & Delay       & Travel time      \\  		  		 		\midrule  		  		 		 		 		 		
%		10  & 0.437        & 2.173       & 2.638        \\ 		 		 		 		 		
%		20   &0.392 &2.144 & 2.574        \\
%		30          & 0.380       & 2.147      & 2.550       \\
%		40 & 0.371  & 2.134 & 2.553 \\ 		 		 		 		 		
%		50      & 0.415        & 2.159       & 2.595         \\		  		 		 		\bottomrule  	  	 	 	 	 	
%	\end{tabular}        
%\caption{Queue Length of RELight with different UTD ratio}        
%\label{tab:booktabs}       
%\end{table}

%\begin{figure}[h]
%	\centering
%	\includegraphics[width=0.8\linewidth]{utd.png}
%	\caption{Queue Length of RELight with different UTD}
%\end{figure} 

\paragraph{N Q-function and M choosed Q-function}

In this part, we conduct comparative experiments on different combinations of N and M, select the appropriate size of N and the optimal number of subsets M. By selecting the M parameter, we can effectively improve the convergence speed, accelerate the convergence process, and does not affect the learning effect.
At the same time, we keep UTD set to 40.
The data set used is also config unequal. According to the figure 3, it is obvious that when we ensemble 10 single DDQN (N=10), the effect of the strategy learned by the agent is much better than that of using only a single DDQN. 
%From the above table,
We can see that when N=10, M takes 4 to get a shorter queue length. This shows that in the ensemble Q functions, we randomly select a subset of the Q function for optimization and learn a better strategy.
\section{CONCLUSION}
In this article, we use the ensemble reinforcement learning method to solve the traffic signal control problem.
During the entire experiment and testing process, no prior knowledge and pre-training is required. This method can learn the optimal strategy only through complete online learning, and a large number of experiments have proved the superiority of our method.
We conducted extensive experiments on a variety of synthetic traffic data and real traffic data and proved that the superiority of our proposed method exceeds the current best method.
In addition, we optimized the RELight algorithm by comparing UTD, M, N, and other parameters to select the optimal parameters for traffic signal control.
%In future work, consideration should be given to considering the joint control of multiple intersections, to be more suitable for actual application scenarios.
%\section*{Acknowledgments}
%
%The preparation of these instructions and the \LaTeX{} and Bib\TeX{}
%files that implement them was supported by Schlumberger Palo Alto
%Research, AT\&T Bell Laboratories, and Morgan Kaufmann Publishers.
%Preparation of the Microsoft Word file was supported by IJCAI.  An
%early version of this document was created by Shirley Jowell and Peter
%F. Patel-Schneider.  It was subsequently modified by Jennifer
%Ballentine and Thomas Dean, Bernhard Nebel, Daniel Pagenstecher,
%Kurt Steinkraus, Toby Walsh and Carles Sierra. The current version
%has been prepared by Marc Pujol-Gonzalez and Francisco Cruz-Mencia.
\section{Acknowledgements} 
The work was supported in part by the National Science Foundation of China grants 61876138.
% Any opinions, findings, and conclusions expressed here are those of the authors anddo not necessarily reflect the views of the funding agencies.

\bibliographystyle{named}
\bibliography{ijcai22}

\end{document}